\documentclass[journal]{IEEEtran}

\usepackage{subcaption}
\usepackage{graphicx}
\usepackage{amsmath}
\usepackage[utf8]{inputenc}

\begin{document}
\title{Survey of Deep Reinforcement Learning for Motion Planning of Autonomous Vehicles}

%\author{Szilárd~Aradi,~\IEEEmembership{Member,~IEEE,}% <-this % stops a space
%\thanks{Szilárd Aradi is with the Department of Control for Transportation and %Vehicle Systems, Budapest University of Technology and Economics, Hungary %e-mail: aradi.szilard@mail.bme.hu}% <-this % stops a space
%\thanks{Manuscript received Month dd, yyyy; revised Month dd, yyyy.}}
\author{Szilárd Aradi$^{1}$
\thanks{$^{1}$ Department of Control for Transportation and Vehicle Systems, Budapest University of Technology and Economics, Hungary\newline e-mail: aradi.szilard@mail.bme.hu}}% <-this % stops a space

\maketitle

\begin{abstract}  
Academic research in the field of autonomous vehicles has reached high popularity in recent years related to several topics as sensor technologies, V2X communications, safety, security, decision making, control, and even legal and standardization rules. Besides classic control design approaches, Artificial Intelligence and Machine Learning methods are present in almost all of these fields. Another part of research focuses on different layers of Motion Planning, such as strategic decisions, trajectory planning, and control. A wide range of techniques in Machine Learning itself have been developed, and this article describes one of these fields, Deep Reinforcement Learning (DRL).
The paper provides insight into the hierarchical motion planning problem and describes the basics of DRL. 
The main elements of designing such a system are the modeling of the environment, the modeling abstractions, the description of the state and the perception models, the appropriate rewarding, and the realization of the underlying neural network.
The paper describes vehicle models, simulation possibilities and computational requirements.
Strategic decisions on different layers and the observation models, e.g., continuous and discrete state representations, grid-based, and camera-based solutions are presented.
The paper surveys the state-of-art solutions systematized by the different tasks and levels of autonomous driving, such as car-following, lane-keeping, trajectory following, merging, or driving in dense traffic.
Finally, open questions and future challenges are discussed.

\end{abstract}


\section{Introduction}

\begin{figure}[htbp]
      \centering
      \includegraphics[width=\linewidth]{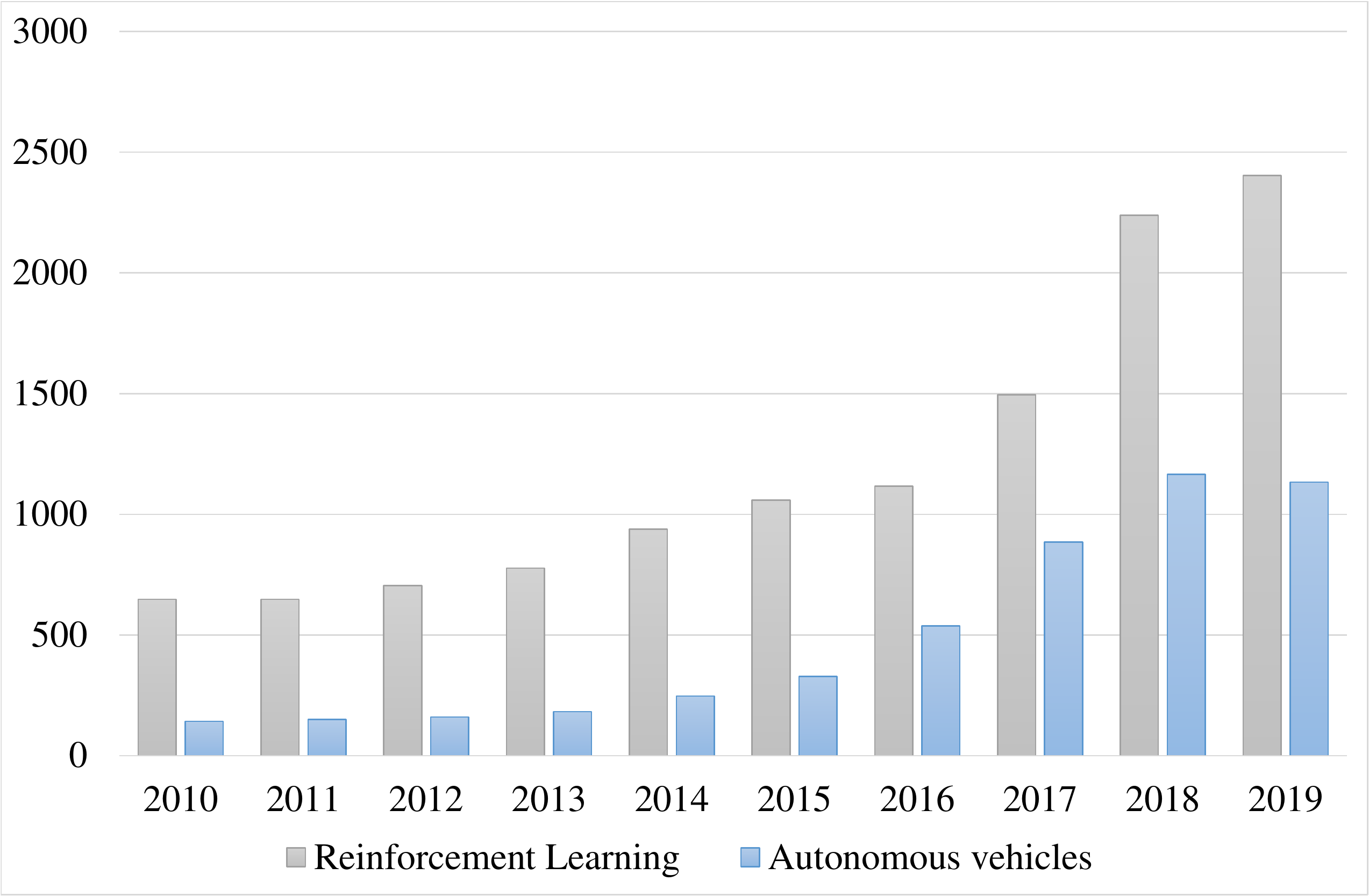}
      \caption{Web of Science topic search for "Deep Reinforcement Learning" and "Autonomous Vehicles (2020.01.17.)"}
\label{Fig:wos}
\end{figure}

Motion planning for autonomous vehicle functions is a vast and long-researched area using a wide variety of approaches such as different optimization techniques, modern control methods, artificial intelligence, and machine learning. This article presents the achievements of the field from recent years focused on Deep Reinforcement Learning (DRL) approach. DRL combines the classic reinforcement learning with deep neural networks, and gained popularity after the breakthrough article from Deepmind \cite{Mnih2013PlayingLearning, Mnih2015Human-levelLearning}. In the number of research papers about autonomous vehicles and the DRL has been increased in the last few years (see Fig. \ref{Fig:wos}.), and because of the complexity of the different motion planning problems, it is a convenient choice to evaluate the applicability of DRL for these problems.

\subsection{The Hierarchical Classification of Motion Planning for Autonomous Driving}
\begin{figure*}[tbhp]
    \centering
    \includegraphics[width=\textwidth]{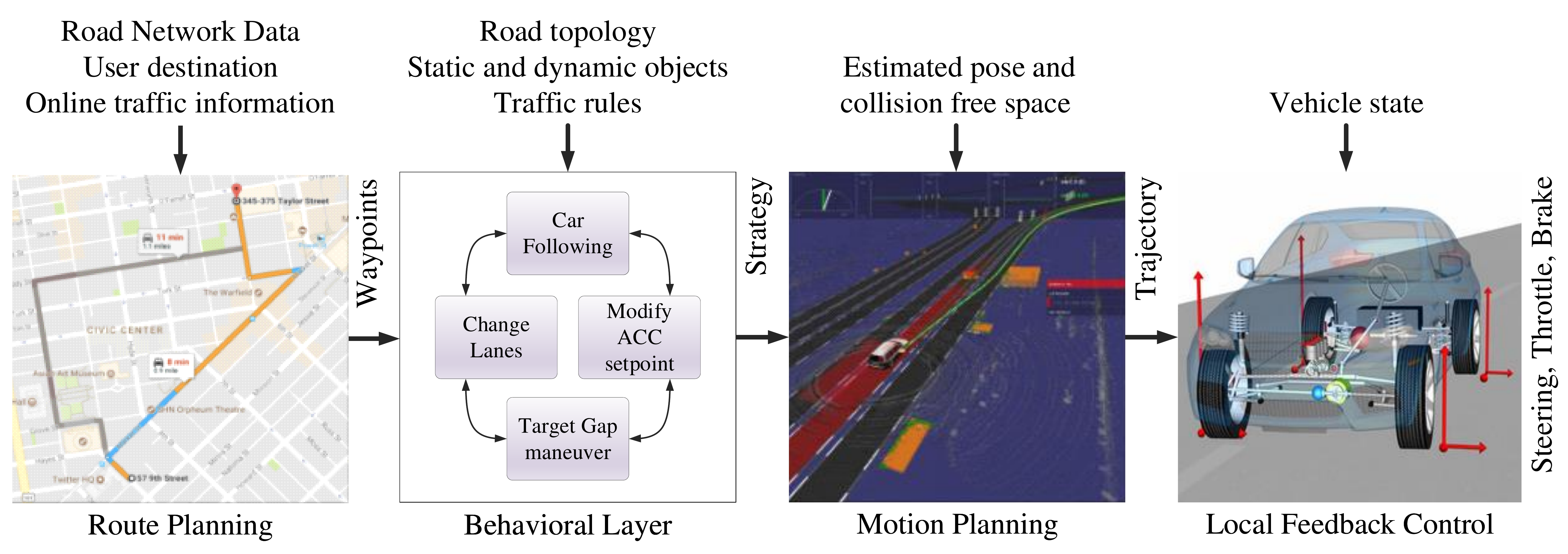} 
    \caption{Layers of motion planning}
    \label{fig:motion_planning}
\end{figure*}

Using deep neural networks for self-driving cars gives the possibility to develop "end-to-end" solutions where the system operates like a human driver: its inputs are the travel destination, the knowledge about the road network and various sensor information, and the output is the direct vehicle control commands, e.g., steering, torque, and brake. However, on the one hand, realizing such a scheme is quite complicated, since it needs to handle all layers of the driving task, on the other hand, the system itself behaves like a black box, which raises design and validation problems. By examining the recent advantages in the field, it can be said that most researches focus on solving some sub-tasks of the hierarchical motion planning problem. This decision-making system of autonomous driving can be decomposed into at least four layers, as stated in \cite{Paden2016} (see Fig.\ref{fig:motion_planning}.). Route planning, as the highest level, defines the way-points of the journey based on the map of the road network, with the possibility of using real-time traffic data. Though optimal route choice has a high interest among the research community, papers dealing with this level do not employ reinforcement learning. A comprehensive study on the subject can be found in \cite{Bast2016RouteNetworks}.

The Behavioral layer is the strategic level of autonomous driving. With the given way-points, the agent decides on the short term policy, by taking into consideration the local road topology, the traffic rules, and the perceived state of other traffic participants. Having a finite set of available actions for the driving context, the realization of this layer is usually a finite state-machine having basic strategies in its states (i.e., car following, lane changing, etc.) with well-defined transitions between them based on the change of the environment. However, even with the full knowledge of the current state of the traffic, the future intentions of the surrounding drivers are unknown, making the problem partially observable \cite{Brechtel2014ProbabilisticPOMDPs}. Hence the future state not only depends on the behavior of the ego vehicle but also relies on unknown processes; this problem forms a Partially Observable Markov Decision Process (POMDP). Different techniques exist to mitigate these effects by predicting the possible trajectories of other road users, like in \cite{Wiest2012ProbabilisticModels}, where the authors used gaussian mixture models, or in \cite{Dou2016LaneClassifiers}, where support vector machines and artificial neural networks were trained based on recorded traffic data. Since finite action POMDPs are the natural way of modeling reinforcement learning problems, a high amount of research papers deal with this level, as can be seen in the sections of the paper.   
To carry out the strategy defined by the behavioral layer, the motion planning layer needs to design a feasible trajectory consisting of the expected speed, yaw, and position states of the vehicle on a short horizon. Naturally, on this level, the vehicle dynamics has to be considered, hence classic exact solutions of motion planning are impractical since they usually assume holonomic dynamics. It has long been known that the numerical complexity of solving the motion planning problem with nonholonomic dynamics is Polynomial-Space Algorithm (PSPACE) \cite{Reif1979ComplexityGeneralizations}, meaning it is hard to elaborate an overall solution by solving the nonlinear programming problem in real-time \cite{Hegedus2018HybridNetworks}. On the other hand, the output representation of the layer makes it hard to directly handle it with "pure" reinforcement learning, only a few papers deal solely with this layer, and they usually use DRL to define splines as a result of the training \cite{Saxena2019DrivingLearning,Feher2019HybridPlanning}.

At the lowest level, the local feedback control is responsible for minimizing the deviation from the prescribed path or trajectory. A significant amount of papers reviewed in this article deals with the aspects of this task, where lane-keeping, trajectory following, or car following is the higher-level strategy. Though at this level, the action space becomes continuous, and classical approaches of RL can not handle this. Hence discretization of the control outputs is needed, or - as in some papers - continuous variants of DRL are used. 

\subsection {Reinforcement Learning}

As an area of Artificial Intelligence and Machine Learning, Reinforcement learning (RL) deals with the problem of a learning agent placed in an environment to achieve a goal. Contrary to supervised learning, where the learner structure gets examples of good and bad behavior, the RL agent must discover by trial and error how to behave to get the most reward \cite{Sutton2017ReinforcementIntroduction}.
For this task, the agent must percept the state of the environment at some level and based on this information, and it needs to take actions that result in a new state. As a result of its action, the agent receives a reward, which aids in the development of future behavior. To ultimately formulate the problem, modeling the state transitions of the environment, based on the actions of the agent is also a necessity. This leads to the formulation of a POMDP defined by the functions of $(\mathcal{S}, \mathcal{A}, T, R, \Omega, O)$, where $\mathcal{S}$ is the set of environment states, $\mathcal{A}$ is the set of possible actions in that particular state, $T$ is the transition function between the states based on the actions, $R$ is the reward for the given $(\mathcal{S}, \mathcal{A})$ pair, while $\Omega$ is the set of observations, and $O$ is the sensor model. The agent in this context can be formulated by any inference model whose parameters can be modified in response to the experience gained. In the context of Deep Reinforcement Learning, this model is implemented by neural networks.

 \begin{figure*}[tbhp]
    \centering
    \includegraphics[width=\textwidth]{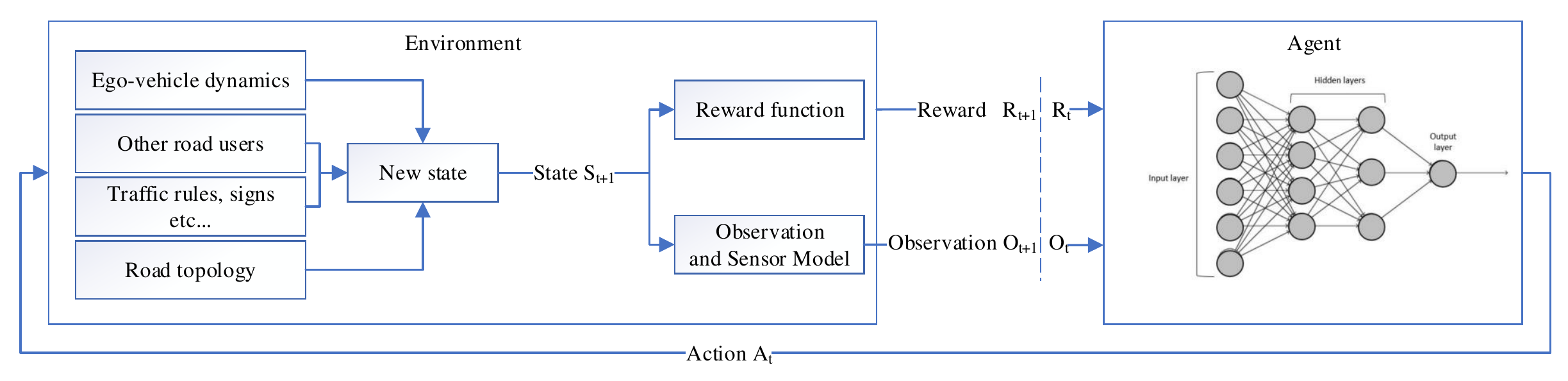} 
    \caption{The POMDP model for Deep Reinforcement Learning based autonomous driving}
    \label{fig:pomdp}
\end{figure*}

The problem in the POMDP scenario is that the current actions affect the future states, therefore the future rewards, meaning that for optimizing the behavior for the cumulative reward throughout the entire episode, the agent needs to have information about the future consequences of its actions. RL has two main approaches for determining the optimal behavior: value-based and policy-based methods. 

The original concept using a value-based method is the Deep-Q Learning Network (DQN) introduced in \cite{Mnih2013PlayingLearning}. Described briefly, the agent predicts a so-called Q value for each state-action pair, which formulate the expected immediate and future reward. From this set, the agent can choose the action with the highest value as an optimal policy or can use the values for exploration during the training process. The main goal is to learn the optimal Q function, represented by a neural network in this case. This can be done by conducting experiments, calculating the discounted rewards of the future states for each action, and updating the network by using the Bellman-equation \cite{Bellman1957DynamicProgramming} as a target.
Using the same network for value evaluation and action selection results in unstable behavior and slow learning in noisy environments. Meta-heuristics, such as experience replay, can handle this problem, while other variants of the original DQN exist, such as Double DQN \cite{VanHasselt2016DeepQ-learning} or Dueling DQN \cite{Wang2015DuelingLearning}, separating the action and the value prediction streams, leading to faster and more stable learning. 

Policy-based methods target at choosing the optimal behavior directly, where the policy $\pi_\Theta$ is a function of $(\mathcal{S}, \mathcal{A})$. Represented by a neural network, with a softmax head, the agent generally predicts a normalized probability of the expected goodness of the actions. In the most natural implementation, this output integrates the exploration property of the RL process. In advanced variants, such as the actor-critic, the agent uses different predictions for the value and the action \cite{Silver2014DeterministicAlgorithms}. Initially, RL algorithms use finite action space, though, for many control problems, they are not suitable. To overcome this issue in \cite{Lillicrap2015ContinuousLearning} introduced the Deep Deterministic Policy Gradients (DDPG) agent, where the actor directly maps states to continuous actions.

For complex problems, the learning process can still be long or even unsuccessful. It can be soluted in many ways:
\begin{itemize}
    \item Curriculum learning describes a type of learning in which the training starts with only easy examples of a task and then gradually increase difficulty. This approach is used in \cite{Qiao2018AutomaticallyEnvironment, Bouton2019Cooperation-AwareTraffic, Kaushik2018OvertakingLearning}.
    \item Adversarial learning aims to fool models through malicious input. Papers using variants of this technique are: \cite{Ferdowsi2018RobustSystems, Ma2018ImprovedLearning}
    \item Model-based action choice, such as the MCTS based solution of Alpha-Go, can reduce the effect of the problem of distant rewarding.
\end{itemize}

Since reinforcement learning models the problem as a POMDP, a discrete-time stochastic control process, the solutions need to provide a mathematical framework for this decision making in situations where outcomes are partly random and partly under the control of a decision-maker, while the states are also partly observable \cite{Kaelbling1998PlanningDomains}. In the case of motion planning for autonomous or highly automated vehicles, the tuple $(\mathcal{S}, \mathcal{A}, T, R, \Omega, O)$ of the POMDP is illustrated in Fig. \ref{fig:pomdp} and can be interpreted as follows:
 
 $\mathcal{S}, \mathcal{A}, T,$ and $R$ describe the MDP, the modeling environment of the learning process. It can vary depending on the goals, though in our case it needs to model the dynamics of the vehicle, the surrounding static and dynamic objects, such as other participants of the traffic, the road topology, lane markings, signs, traffic rules, etc. $\mathcal{S}$ holds the current actual state of the simulation. $A$ is the possible set of actions of the agent driving the ego-car, while $T$, the so-called state-transition function updates the vehicle state and also the states of the traffic participants depending on the action of the vehicle. The different levels of abstraction are described in section \ref{ss_model}. Many research papers use different software platforms for modeling the environment. A brief collection of the used frameworks are presented in section \ref{ss_simulators}. $R$ is the reward function of the MDP, section \ref{ss_rewarding} gives a summary on this topic. 
 
$\Omega$ is the set of observations the agent can experience in the world, while $O$ is the observation function giving a possibility distribution over the possible observations. In more uncomplicated cases, the studies assume full observability and formulate the problem as an MDP, though in many cases, the vehicle does not possess all information. Another interesting topic is the representation of the state observation, which is a crucial factor for the architecture choice and performance of Deep RL agents. The observation models used in the literature are summarized in section \ref{ss_observation}.  
 
\section{Modeling for Reinforcement Learning}
 
\subsection{Vehicle modeling}\label{ss_model}

Modeling the movement of the ego-vehicle is a crucial part of the training process since it raises the trade-off problem between model accuracy and computational resource. Since RL techniques use a massive number of episodes for determining optimal policy, the step time of the environment, which highly depends on the evaluation time of the vehicle dynamics model, profoundly affects training time. Therefore during environment design, one needs to choose from the simplest kinematic model to more sophisticated dynamics models ranging from 2 Degree of Freedom (2DoF) lateral model to the more and more complex models with a higher number of parameters and complicated tire models.

At rigid kinematic single-track vehicle models, which neglect tire slip and skip, lateral motion is only affected by the geometric parameters. Therefore, they are usually limited to low-speed applications. More details about the model can be found in \cite{Kong2015KinematicDesign}. 
The simplest dynamic models with longitudinal and lateral movements are based on the 3 Degrees of Freedom (3DoF) dynamic bicycle model, usually with a linear tire model. They consider $(V_x, V_y, \dot{\Psi})$ as independent variables, namely longitudinal and lateral speed, and yaw rate.
A more complex model is the four-tire 9 Degrees of Freedom (9DoF) vehicle model, where amongst the parameters of the 3DoF, body roll and pitch $(\dot{\Theta}, \dot{\Phi})$ and the angular velocities of the four wheels $({\omega_{fl},\omega_{fr},\omega_{rl},\omega_{rr}})$ are also considered, to calculate tire forces more precisely. Hence the model takes into account both the coupling of longitudinal and lateral slips and the load transfer between tires. 

Though the kinematic model seems quite simplified, and as stated in \cite{Polack2017TheVehicles}, such a model can behave significantly different from an actual vehicle, though for the many control situations, the accuracy is suitable \cite{Kong2015KinematicDesign}.

According to \cite{Polack2017TheVehicles}, using a kinematic bicycle model with a limitation on the lateral acceleration at around $0.5g$ or less provides appropriate results, but only with the assumption of dry road. Above this limit, the model is unable to handle dynamics. Hence a more accurate vehicle model should be used when dealing with higher accelerations to push the vehicle’s dynamics near its handling limits. 

Regarding calculation time, based on the kinematic model, the calculation of the 3DoF model can be $10\dots 50$ times higher, and the precise calculation of a 9DoF model with nonlinear tire model can be $100\dots 300$ times higher, which is the main reason for the RL community to use a low level of abstraction.

Modeling traffic and surrounding vehicles is often performed by using unique simulators, as described in section \ref{ss_simulators}. Some authors develop their environments, using cellular automata models  \cite{You2019AdvancedLearning}. Some use MOBIL, which is a general model (minimizing overall braking induced by lane change) to derive lane-changing rules for discretionary and mandatory lane changes for a broad class of car-following models \cite{Kesting2007GeneralModels};  the  Intelligent Driving Model (IDM), a continuous microscopic single-lane model \cite{Treiber2000CongestedSimulations}. 

\subsection{Simulators}
\label{ss_simulators}

Some authors create self-made environments to achieve full control over the model, though there are commercial and Open-source environments that can provide this feature. This section briefly identifies some of them used in recent researches in motion planning with RL. 
   
In modeling the traffic environment, the most popular choice is SUMO (Simulation of Urban MObility), which is a microscopic, inter- and multi-modal, space-continuous and time-discrete traffic flow simulation platform \cite{Krajzewicz2012RecentMObility}. It can convert networks from other traffic simulators such as VISUM, Vissim, or MATSim and also reads other standard digital road network formats, such as OpenStreetMap or OpenDRIVE. It also provides interfaces to several environments, such as python, Matlab, .Net, C++, etc. Though the abstraction level, in this case, is microscopic, and vehicle behavior is limited, its ease of use and high speed makes it an excellent choice for training agents to handle traffic, though it does not provide any sensor model besides the ground truth state of the vehicles. 

Another popular microscopic simulator that has been used commercially and for research also is VISSIM \cite{Fellendorf2010MicroscopicVISSIM}. In  \cite{Ye2019AutomatedEnvironment} it is used for developing car-following behavior and lane changing decisions.

Considering only vehicle dynamics, the most popular choice is TORCS (The Open Racing Car Simulator), which is a modern, modular, highly portable multi-player, multi-agent car simulator. Its high degree of modularity and portability render it ideal for artificial intelligence research \cite{Wymann2014TORCS:Simulator}. Interfacing with python, the most popular AI research environment is comfortable and runs at an acceptable speed. TORCS also comes with different tracks, competing robots, and several sensor models.

It is assumed that for vehicle dynamics, the best choices would be the professional tools, such as CarSIM \cite{CarSIMCorporation} or CarMaker \cite{CarMakerAutomotive}, though the utilization of these softwares can not be found in the reinforcement learning literature. This may be caused by the fact that these are expensive commercial platforms, though more importantly, their lack of python interfaces and high precision, but resource-intensive models prevent them from running several episodes within a reasonable time.   

For more detailed sensor models or traffic, the authors usually use Airsim, Udacity Gazebo/ROS, and CARLA:

AirSim, used by a recent research in \cite{An2019Decision-MakingDriving}, is a simulator initially developed for drones built on Unreal Engine now has a vehicle extension with different weather conditions and scenarios. 

Udacity, used in \cite{Wang2019LaneConstraints}, is a  simulator that was built for Udacity's Self-Driving Car Nanodegree \cite{WelcomeSimulator} provides various sensors, such as high quality rendered camera image LIDAR and Infrared information, and also has capabilities to model other traffic participants.
   
Another notable mention is CARLA, an open-source simulator for autonomous driving research. CARLA has been developed from the ground up to support the development, training, and validation of autonomous urban driving systems. In addition to open-source code and protocols, CARLA provides open digital assets
(urban layouts, buildings, vehicles) that were created for this purpose and can be used freely. The simulation platform supports flexible specification of sensor
suites and environmental conditions \cite{Dosovitskiy2017CARLA:Simulator}. 

Though this section provides only a brief description of the simulators, a more systematic review of the topic can be found in \cite{Rosique2019AResearch}.

\subsection{Action Space}\label{ss_action}
The choice of action space highly depends on the vehicle model and task designed for the reinforcement learning problem in each previous research. Though two main levels of control can be found: one is the direct control of the vehicle by steering braking and accelerating commands, and the other acts on the behavioral layer and defines choices on strategic levels, such as lane change, lane keeping, setting ACC reference point, etc. At this level, the agent gives a command to low-level controllers, which calculate the actual trajectory. Only a few papers deal with the motion planning layer, where the task defines the endpoints $(x,y,\theta)$, and the agent defines the knots of the trajectory to follow represented as a spline, as can be seen in \cite{Feher2019HybridPlanning}. Also, few papers deviate from vehicle motion restrictions and generate actions by stepping in a grid, like in classic cellular automata microscopic models \cite{Kashihara2017DeepJunction}.

Some papers combine the control and behavioral layers by separating longitudinal and lateral tasks, where longitudinal acceleration is a direct command, while lane changing is a strategic decision like in \cite{Nageshrao2019AutonomousLearning}.

The behavioral layer usually holds a few distinct choices, from which the underlying neural network needs to choose, making it a classic reinforcement learning task with finite actions.

Though on the level of control, the actuation of vehicles, i.e., steering, throttle, and braking, are continuous parameters and many reinforcement learning techniques like DQN and PG can not handle this since they need finite action set, while some, like DDPG, works with continuous action space. 
To adapt to the finite action requirements of the RL technique used, most papers discretizes the steering and acceleration commands to 3 to 9 possibilities per channel. The low number of possible choices pushes the solution farther from reality, which could raise vehicle dynamics issues with uncontrollable slips, massive jerk, and yaw-rate, though the utilization of kinematic models sometimes covers this in the papers. A large number of discrete choices, however, ends up in an exponential growth in the possible outcomes in the POMDP approach, which slows down the learning process.

\subsection{Rewarding}
\label{ss_rewarding}
During training, the agent tries to fulfill a task, generally consisting of more than one step. This task is called an episode. An episode ends if one of the following conditions are met:
\begin{itemize}
    \item The agent successfully fulfills the task;
    \item The episode reaches a previously defined steps
    \item A terminating condition rises.
\end{itemize}
The first two cases are trivial and depend on the design of the actual problem. Terminal conditions are typically situations where the agent reaches a state from which the actual task is impossible to fulfill, or the agent makes a mistake that is not acceptable. Vehicle motion planning agents usually use terminating conditions, such as:
collision with other participants or obstacles or leaving the track or lane, since these two inevitably end the episode. There are lighter approaches, where the episode terminates with failure before the accident occurred, with examples of having a too high tangent angle to the track or reaching too close to other participants. These "before accident" terminations speed up the training by bringing the information of failure forward in time, though their design needs caution \cite{Alizadeh2019AutomatedEnvironment}. 
    
Rewarding plays the role of evaluating the goodness of the choices the agent made during the episode giving feedback to improve the policy. The first important aspect is the timing of the reward, where the designer of the reinforcement learning solution needs to choose a mixture of the following strategies all having their pros and cons:
\begin{itemize}
    \item Giving reward only at the end of the episode and discounting it back to the previous $(\mathcal{S}, \mathcal{A})$ pairs, which could result in a slower learning process, though minimizes the human-driven shaping of the policy. 
    \item Giving immediate reward at each step by evaluating the current state, naturally discount also appears in this solution, which results in significantly faster learning, though the choice of the immediate reward highly affects the established strategy, which sometimes prevents the agent from developing better overall solutions than the one that gave the intention of the designed reward.
    \item An intermediate solution can be to give a reward in predefined periods or travel distance \cite{Feher2018Q-learningKeeping}, or when a good or bad decision occurs. 
\end{itemize}

In the area of motion planning, the end episode rewards are calculated from the fulfillment or failure of the driving task. The overall performance factors are generally: time of finishing the task, keeping the desired speed or achieving as high average speed as possible, yaw or distance from lane middle or the desired trajectory, overtaking more vehicles, achieve as few lane changes as possible \cite{Bai2019DeepTraffic}, keeping right \cite{Wolf2018AdaptiveStates, Aradi2018PolicyDriving} etc.   
Rewarding systems also can represent passenger comfort, where the smoothness of the vehicle dynamics is enforced. The most used quantitative measures are the longitudinal acceleration \cite{Xu2018AHighways}, lateral acceleration \cite{Wang2018AManeuvers, Ronecker2019DeepDriving} and jerk \cite{Zhu2019SafeDriving, Saxena2019DrivingLearning}.

In some researches, the reward is based on the deviation from a  dataset \cite{Zhu2018Human-likeLearning}, or calculated as a deviation from a reference model like in \cite{Hoel2018AutomatedLearning}. These approaches can provide favorable results, though a bit tends from the original philosophy of reinforcement learning since a previously known strategy could guide the learning.  

\subsection{Observation Space}
\label{ss_observation}

The observation space describes the world to the agent. It needs to give sufficient information for choosing the appropriate action, hence - depending on the task - it contains the following knowledge:
\begin{itemize}
    \item The state of the vehicle in the world, e.g., position, speed, yaw, etc.
    \item Topological information like lanes, signs, rules, etc.
    \item Other participants: surrounding vehicles, obstacles, etc.
\end{itemize}

The reference frame of the observation can be absolute and fixed to the coordinate system of the world, though as the decision process focuses on the ego-vehicle, it is more straightforward to choose an ego-centric reference frame pinned to the vehicle's coordinate system, or the vehicle's position in the world, and the orientation of the road. It allows concentrating the distribution of visited states around the origin in both position, heading, and velocity space, as other vehicles are often close to the ego-vehicle and with similar speed and heading, reducing the region of state-space in which the policy must perform. \cite{Leurent2018}

\subsubsection{Vehicle state observation}
\begin{figure}[htpb]
    \centering
    \includegraphics[width=\linewidth]{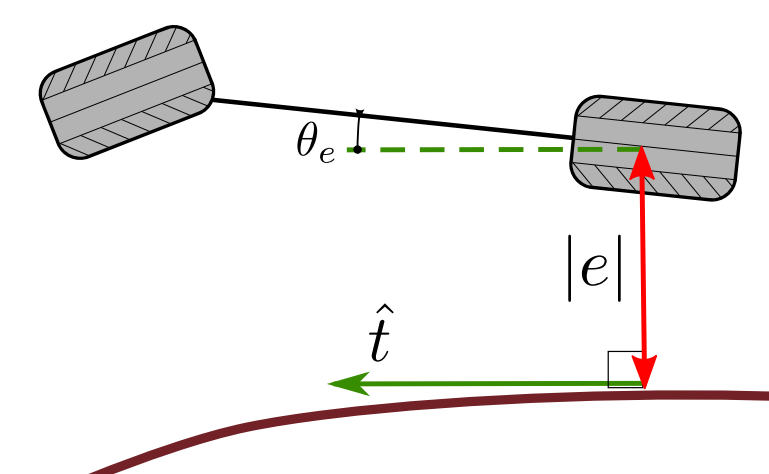} 
    \caption{Observation for basic vehicle state (source: \cite{Paden2016})}
    \label{fig:lane_keeping_state}
\end{figure}

For lane keeping, navigation, simple racing, overtaking, or maneuvering tasks, the most commonly used and also the simplest observation for the ego vehicle consists of the continuous variables of $(|e|, v, \theta_e)$ describing the lateral position from the center-line of the lane, vehicle speed, and yaw angle respectively. (see Fig. \ref{fig:lane_keeping_state}).
This information is the absolute minimum for guiding car-like vehicles, and only eligible for the control of the classical kinematic car-like models, where the system implies the motion without skidding assumption. Though in many cases in the literature, this can be sufficient, since the vehicles remain deep in the dynamically stable region. 

For tasks, where more complex vehicle dynamics is inevitable, such as racing situations, or where the stability of the vehicle is essential, this set of observable state would not be enough, and it should be extended with yaw, pitch, roll, tire dynamics, and slip.
%It is difficult to find studies, that focus on state of the reinforcement learning for vehicle motion planning tasks. 

\subsubsection{Environment observation}
Getting information about the surroundings of the vehicle and representing it to the learning agent shows high diversity in the literature. Different levels of sensor abstractions can be observed:

\begin{itemize}
\item sensor level, where camera images, lidar or radar information is passed to the agent;
\item intermediate level, where idealized sensor information is provided;
\item ground truth level, where all detectable and non-detectable information is given. 
\end{itemize}

The structure of the sensor model also affects the neural network structure of the Deep RL agent since image like, or array-like inputs infer 2D or 1D CNN structures, while the simple set of scalar information results in a simple dense network. There are cases where these two kinds of inputs are mixed. Hence the network needs to have two different types of input layers.  

Image-based solutions usually use front-facing camera images extracted from 3D simulators to represent the observation space. The data is structured in a ($C$ x $W$ x $H$) sized matrix, where $C$ is the number of channels, usually one for intensity images and three for RGB, while $W$ and $H$ are the width and height resolution of the image. In some cases, for the detection of movement, multiple images are fed to the network in parallel.
Sometimes it is convenient to down-sample the images - like ($1$x$48$x$27$) in \cite{Wolf2017LearningQ-Networks} or ($3$x$84$x$84$) in \cite{Jaritz2018End-to-EndLearning, Perot2017End-to-EndLearning} - for data and network compression purposes. Since images hold the information in an unstructured manner, i.e., the state information, such as object positions, or lane information are deeply encoded in the data, deep neural networks, such as CNN, usually need large samples and time to converge \cite{Li2019ReinforcementNotes}. This problem escalates, with the high number of steps that the RL process requires, resulting in a lengthy learning process, like $1.5M$ steps in \cite{Wolf2017LearningQ-Networks} or $100M$ steps in \cite{Jaritz2018End-to-EndLearning}.    

Many image-based solutions propose some kind of preprocessing of the data to overcome this issue. In \cite{Li2019ReinforcementNotes}, the authors propose a framework for vision-based lateral control, which combines DL and RL methods. To improve the perception accuracy, an MTL (Multitask learning) CNN model is proposed to learn the critical track features, which are used to locate the vehicle in the track coordinate, and trains a policy gradient RL controller to solve the continuous sequential decision-making problem. Naturally, this approach can also be viewed as an RL solution with structured features, though the combined approach has its place in the image-based solutions also.  

\begin{figure}[htbp]
   \centering
   \includegraphics[width=\linewidth]{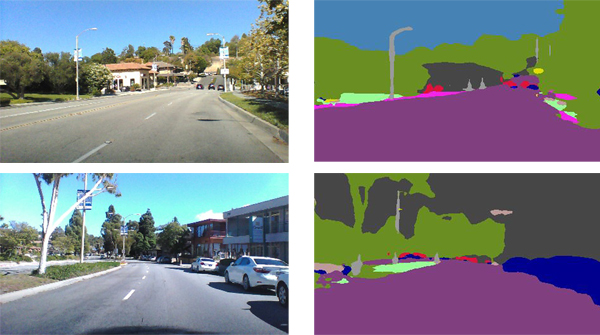}
   \caption{ Real images from the driving data and their semantic segmentations (source:\cite{Xu2018AutonomousTranslation})}
   \label{Fig:segmentedimage}
\end{figure}

Another approach could be the simplification of the unstructured data. In \cite{Kotyan2019SelfAgent} Kotyan et al. uses the difference image as the background subtraction between the two consecutive frames as an input, assuming this image contains the motion of the foreground and the underlying neural network would focus more on the features of the foreground than the background. By using the same training algorithm, their results showed that the including difference image instead of the original unprocessed input needs approximately 10 times less training steps to achieve the same performance.   
The second possibility is, instead of using the original image as an input, it can be driven through an image semantic segmentation network as proposed in \cite{Xu2018AutonomousTranslation}. As the authors state: "Semantic image contains less information compared to the original image, but includes most information needed by the agent to take actions. In other words, semantic image neglects useless
information in the original image." Another advantage of this approach is that the trained agent can use the segmented output of images obtained from real-world scenarios, since on this level, the difference is much smaller between the simulated and real-world data than in the case of the simulated and real-world images. Fig. \ref{Fig:segmentedimage} shows the $640x400$ resolution inputs used in this research.   

2D or 3D Lidar like sensor models are not common among the recent studies, though they could provide excellent depth-map like information about the environment. Though the same problem arises as with the camera images, that the provided data - let them be a vector for 2D, and a matrix for 3D Lidars - is unstructured. The usage of this type of input only can be found in \cite{Lee2017AutonomousQ-learning}, where the observation emulates a 2D Lidar that provides the distance from obstacles in $31$ directions within the field-of-view of $150^{\circ}$, and agent uses sensor data as its state.  A similar input structure, though not modeling a Lidar, since there is no reflection, which is provided by TORCS and used in \cite{Kaushik2018OvertakingLearning}, is to represent the lane markings with imagined beam sensors. The agent in the cited example uses readings from 19 sensors with a 200m range, presenting at every $10^\circ$ on the front half of
the car returning distance to the track edge.

\begin{figure}[htbp]
\begin{subfigure}{0.31\linewidth}
\includegraphics[width=\linewidth]{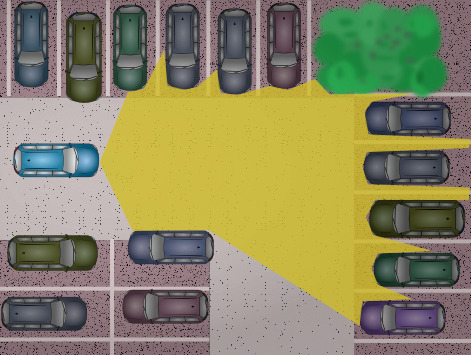}
\caption{Sensors} 
\end{subfigure}
\hspace*{\fill} % separation between the subfigures
\begin{subfigure}{0.31\linewidth}
\includegraphics[width=\linewidth]{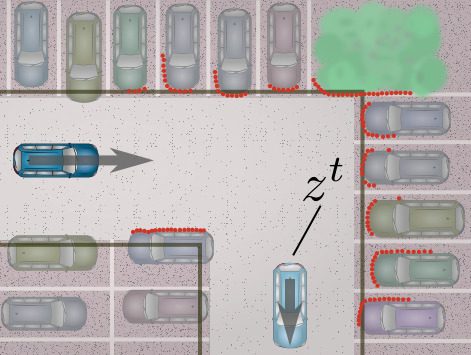}
\caption{Target state $z^t$} 
\end{subfigure}
\hspace*{\fill} % separation between the subfigures
\begin{subfigure}{0.31\linewidth}
\includegraphics[width=\linewidth]{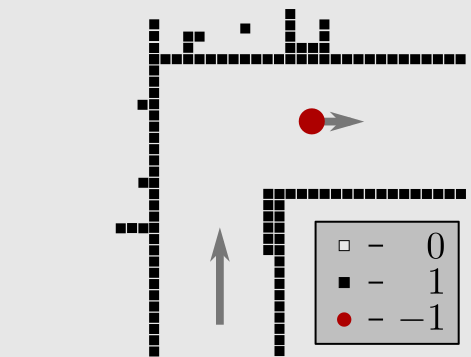}
\caption{Perception Ø} 
\end{subfigure}
\caption{The surrounding from the perspective of the vehicle can be described by a coarse perception map where the target is represented by a red dot (c) (source: \cite{Folkers2019ControllingLearning})} \label{fig:occupancygrid}
\end{figure}

Grid-based path planning methods, like the A* or various SLAM (Simultaneous Localization and Mapping) algorithms exist and are used widespread in the area of mobile robot navigation, where the environment is represented as a spatial map \cite{Elfes1989UsingNavigation}, usually formulated as a 2D matrix assigning to each 2D location in a surface grid one of three possible values: Occupied, free, and unknown \cite{Thrun2006Stanley:Challenge}. This approach can also be used representing probabilistic maneuvers of surrounding vehicles \cite{Deo2018Multi-ModalLSTMs}, or by generating spatiotemporal map from a predicted sequence of movements, motion planning in a dynamic environment can also be achieved \cite{Hegedus2019Graph-basedVehicles}. Though the previously cited examples didn't use RL techniques, they prove that grid representation holds high potential in this field. Navigation in a static environment by using a grid map as the observation, together with position and yaw of the vehicle with an RL agent, is presented in \cite{Folkers2019ControllingLearning} (See Fig.\ref{fig:occupancygrid}). Grid maps are also unstructured data, and their complexity is similar to the semantically segmented images, since the cells store class information in both cases, and hence their optimal handling is using the CNN architecture.
 
\begin{figure}[htbp]
\centering
\begin{subfigure}{0.9\linewidth}
\includegraphics[width=\linewidth]{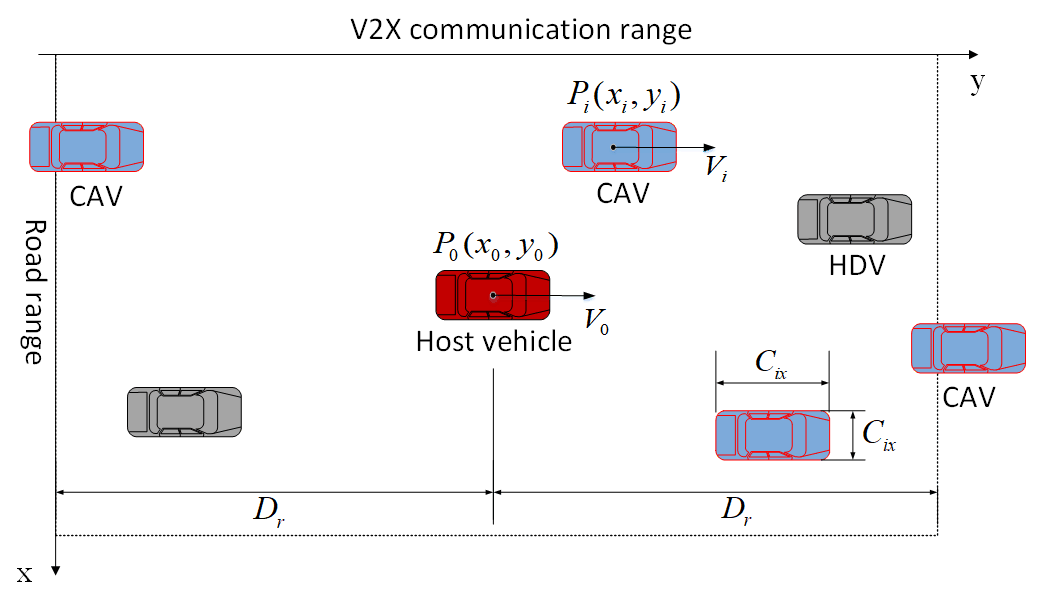}
\caption{Mathematical model for the traffic} 
\end{subfigure}
\hspace*{\fill} % separation between the subfigures
\begin{subfigure}{0.9\linewidth}
\includegraphics[width=\linewidth]{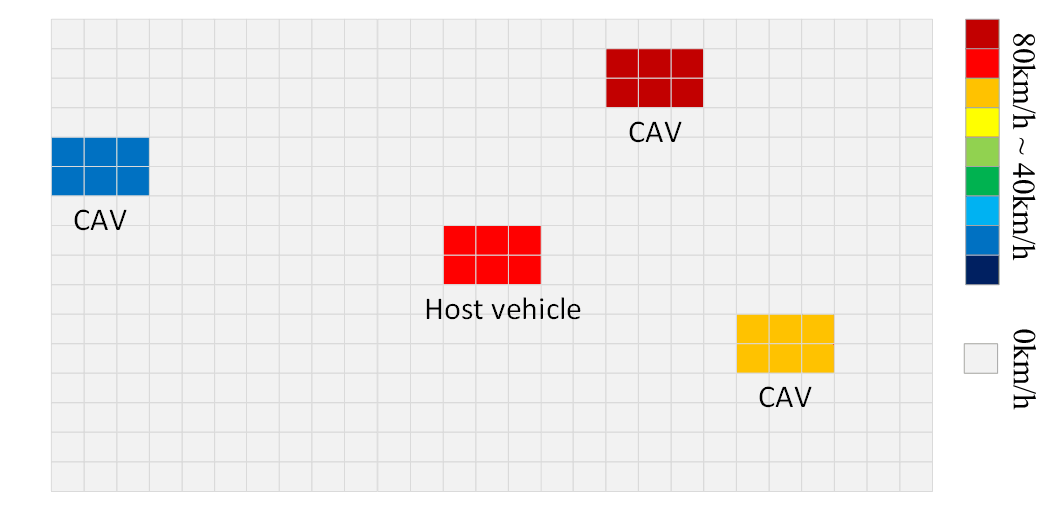}
\caption{Visualization of the Hyper Grid Matrix} 
\end{subfigure}
\hspace*{\fill} % separation between the subfigures

\caption{The visualization of the HDM mapping process (source:\cite{Bai2019DeepTraffic})} \label{Fig:trafficgrid}
\end{figure}

Representing moving objects, i.e. surrounding vehicles in a grid needs not only occupancy, but other information hence the spatial grid's cell need to hold additional information. In \cite{Bai2019DeepTraffic} the authors used equidistant grid, where the ego-vehicle is placed in the center, and the cells occupied by other vehicles represented the longitudinal velocity of the corresponding car (See Fig. \ref{Fig:trafficgrid}). The same approach can also be found in \cite{Ronecker2019DeepDriving}. Naturally this simple representation can not provide information about the lateral movement of the other traffic participants, though they give significantly more than the simple occupancy based ones. 
Equidistant grids are a logical choice for generic environments, where the moving directions of the mobile robot are free, though, in the case of road vehicles, the vehicle mainly follows the direction of the traffic flow. In this case, the spatial representation could be chosen fixed to the road topology, namely the lanes of the road, regardless of its curvature or width. In these lane-based grid solutions, the grid representing the highway has as many rows as the actual lane count, and the lanes are discretized longitudinally.   
The simplest utilization of this approach can be found in \cite{You2019AdvancedLearning}, where the length of the cells is equivalent to the unit vehicle length, and also, the behavior of the traffic acts similar to the classic cellular automata-based microscopic models \cite{Esser1997MicroscopicAutomata}.

This representation, similarly to the equidistant ones, can be used for occupancy, though they still do not hold any information on vehicle dynamics.
\cite{Wang2019CooperativeLearning} is to fed multiple consecutive traffic snapshots into the underlying CNN structure, which inherently extracts the velocity of the moving objects. Representing speed in grid cells is also possible in this setup, for that example can be found in \cite{Wang2019LaneConstraints}, where the authors converted the traffic extracted from the Udacity simulator to the lane-based grid.

\begin{figure}[thpb]
   \centering
   \includegraphics[width=\linewidth]{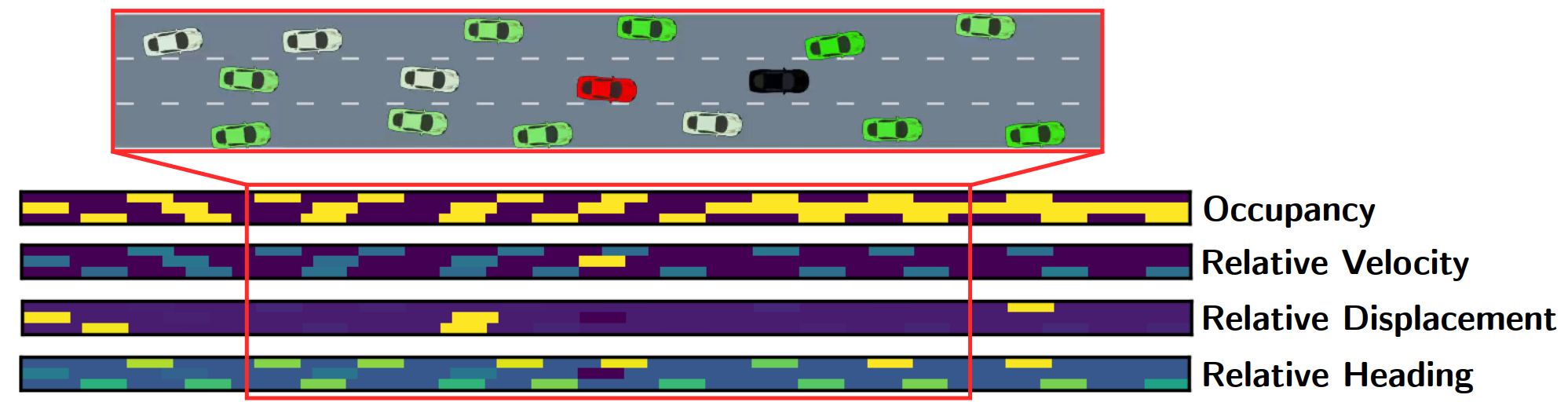}
   \caption{ The simulator state (top, zoomed in) gets converted to a 4 x 3 x (2 x FoV + 1) input observation tensor (bottom) (source:\cite{Saxena2019DrivingLearning})}
   \label{Fig:lanebasedgrid}
\end{figure}

Besides the position and the longitudinal speed of the surrounding vehicles are essential from the aspect of the decision making, other features (such as heading, acceleration, lateral speed) should be considered. Multi-layer grid maps could be used for each vital parameter to overcome this issue. In \cite{Saxena2019DrivingLearning} the authors processed the simulator state to calculate an observation tensor of size 4 x 3 x (2 x FoV + 1), where Fov stands for Field of View and represents the maximum distance of the observation in cell count. There is one channel (first dimension) each for on-road occupancy, relative velocities of vehicles, relative lateral displacements, and relative headings to the ego-vehicle. Fig.\ref{Fig:lanebasedgrid} shows an example of the simulator state and corresponding input observation used for their network.  
  
\begin{figure}[htbp]
      \centering
      \includegraphics[width=0.6\linewidth]{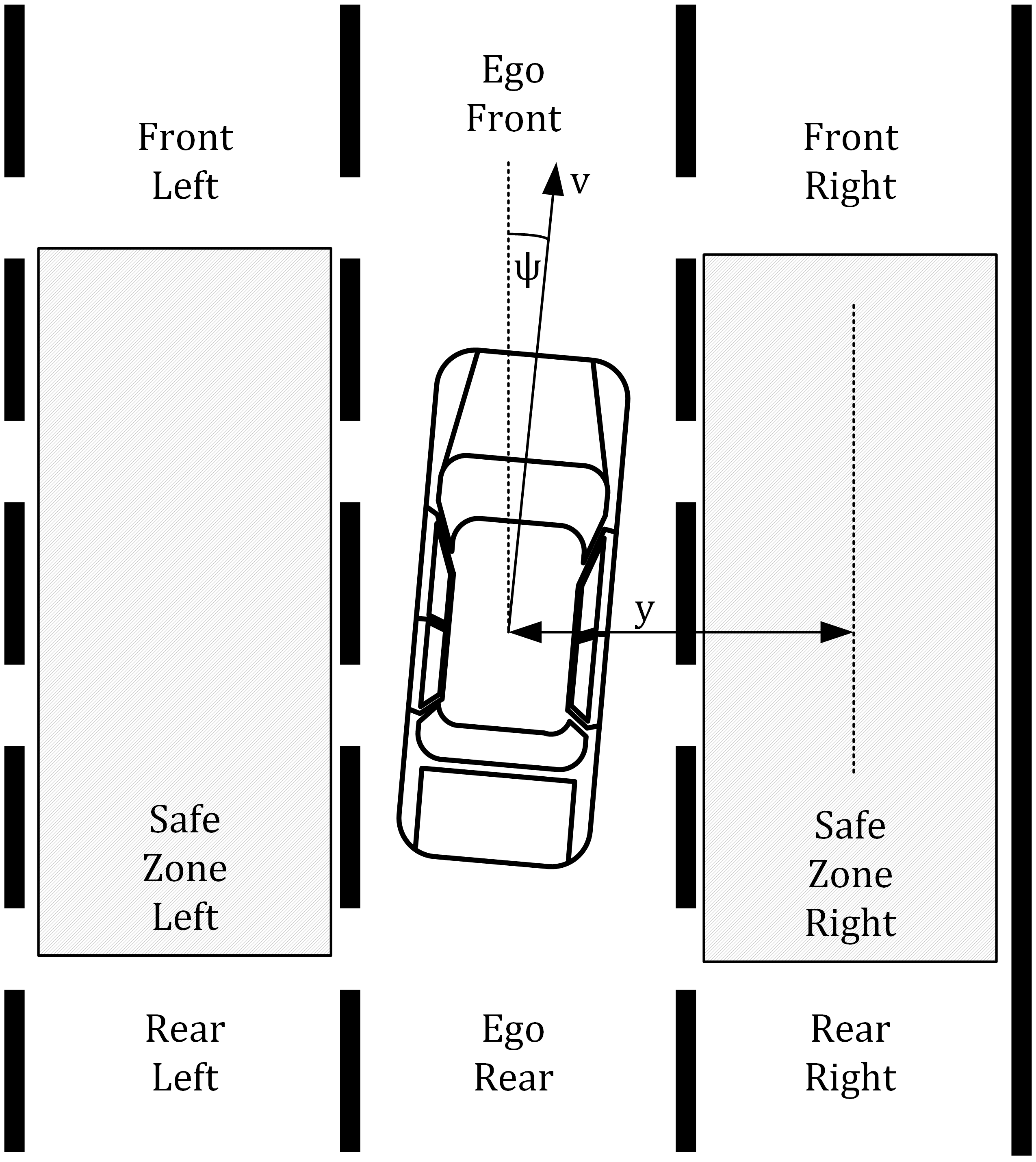}
      \caption{Environment state on the highway \cite{Becsi2018HighwayLearning}}
\label{Fig:aradihighway}
\end{figure}

The previous observation models (image, lidar, or grid-based) all have some common properties: All of them are unstructured datasets, need a CNN architecture to process, which hardens the learning process since the agent simultaneously needs to extract the exciting features and form the policy for action. It would be obvious to pre-process the unstructured data and feed structured information to the agents' network. Structured data refers to any data that resides in a fixed field within a record or file. As an example, for navigating in traffic, based on the task, the parameters of the surrounding vehicles are represented on the same element of the input. In the simplest scenario of car following, the agent only focuses on the leading vehicle, and the input beside the state of the ego vehicle consists of $(d,v)$ as in \cite{Zhu2018Human-likeLearning} or $(d,v, a)$ as in \cite{Shi2019DrivingLearning}, where these parameters are the headway distance, speed, and acceleration of the leading vehicle. Contrary to the unstructured data, these approaches significantly reduce the amount of the input and can be handled with simple DNN structures, which profoundly affects the convergence of the agent's performance. 
For navigating in traffic, i.e., performing merging or lane changing maneuvers, not only the leading vehicle's, but the other surrounding vehicles' states also need to be considered. 
In a merging scenario, the most crucial information is the relative longitudinal position and speed $2 $x$ (dx, dv)$ of the two vehicles bounding the target gap, as used by  \cite{Wang2017FormulationMerge}. Naturally, this is the absolute minimal representation of such a problem, but in the future, more sophisticated representations would be developed.
In highway maneuvering situations, both ego-lane, and neighboring lane vehicles need to be considered, in \cite{Nageshrao2019AutonomousLearning} the authors used the above mentioned $6 $x$ (dx, dv)$ scalar vector is used for the front and rear vehicles in the three interesting lanes. While in \cite{Becsi2018HighwayLearning} the authors extended this information with the occupancy of the neighboring lanes right at the side of the ego-vehicle (See Fig. \ref{Fig:aradihighway}). The same approach can be seen in \cite{Alizadeh2019AutomatedEnvironment}, though extending the number of traced objects to nine.  These researches lack lateral information, though, in \cite{Nageshrao2019AutonomousLearning}, the lateral positions and speeds are also involved in the input vector resulting in a $6$x$ (dx, dy, dvx, dvy)$ structure, logically representing longitudinal and lateral distance, and speed differences to the ego, respectively.
In a special case of handling unsignalized intersection \cite{Bouton2019ReinforcementDriving} the authors also used this formulation scheme where the other vehicle's Cartesian coordinates, speed and heading were considered.

\section{Scenario-based Classification of the Approaches}

Though this survey focuses on Deep Reinforcement Learning based motion planning research, it is essential to mention that some papers try to solve some subtasks of automated driving through classic reinforcement techniques.
One problem of these classic methods, that they can not handle unstructured data, such as images, mid-level radar, or lidar sensing.

The other problem comes from the need of maintaining the Q-table for all $(\mathcal{S},\mathcal{A})$ state-action pairs. This results in space complexity explosion, since the size of the table equals the product of the size of all classes both in state and action. As an example, the Q-learning made in \cite{Loiacono2010LearningLearning} is presented. The authors trained an agent in TORCS, which tries to achieve a policy for the best overtaking maneuver, by taking advantage of the aerodynamic drag. There are only two participants in the scenario, the overtaking vehicle, and the vehicle in front on a long straight track. The state representation contains the longitudinal and lateral distance of the two vehicles and also the the lateral position of the ego-vehicle and the speed difference of the two. 
\begin{table}[htbp]
\centering
\caption{State representation discretization in \cite{Loiacono2010LearningLearning}}
\label{tab:torcsclassic}
\begin{tabular}{l l l}
\hline
Name        & Size           & Class bounds            \\ \hline
$dist_y[m]$     & 6    & \begin{tabular}[c]{@{}l@{}}\{0, 10, 20 ,30, 50, 100, 200\}\end{tabular}                                      \\ 
$dist_x[m]$     & 10   & \begin{tabular}[c]{@{}l@{}}\{-25, -15, -5, -3 , -1, 0, 1, 3, 5, 15, 25\}\end{tabular}    \\ 
$pos[m]$         & 8   & \begin{tabular}[c]{@{}l@{}}\{-10, -5, -2, -1, 0, 1, 2, 5, 10\}\end{tabular}                         \\ 
$\Delta speed[km/h]$ & 9 & \begin{tabular}[c]{@{}l@{}}\{-300, 0, 30, 60, 90, 120,\\ 150, 200, 250, 300\}\end{tabular} \\ \hline
\end{tabular}
\end{table}
The authors discretized this state space to classes of size $(6, 10, 8, 9)$ respectfully (see table \ref{tab:torcsclassic}); and used the minimal lateral action set size of 3, where the actions are sweeping $1m$ to the left or right and maintaining lateral position. Together, this problem generates a Q-table with $6*10*8*9*3 = 12960$ elements. Though a table of this size can be easily handled nowadays, it is easy to imagine that with more complex  problems with more vehicles, more sensors, complex dynamics, denser state and action representation, the table can grow to enormous size. 
A possible reduction is the utilization of the Multiple-Goal Reinforcement Learning Method and dividing the overall problem to sub-tasks, as can be seen in \cite{Ngai2007AutomatedFramework} for overtaking maneuver. In a latter research, the authors widened the problem and separated the driving problem to the tasks of collision avoidance, target seeking, lane following, Lane choice, speed keeping, and steady steering \cite{Ngai2011AManeuvers}.
To reduce problem size, the authors of \cite{Desjardins2011CooperativeApproach} used strategic-level decisions to set movement targets for the vehicles concerning the surrounding ones, and left the low-level control to classic solutions, which significantly reduced the action space.

\begin{figure}[htbp]
      \centering
      \includegraphics[width=0.8\linewidth]{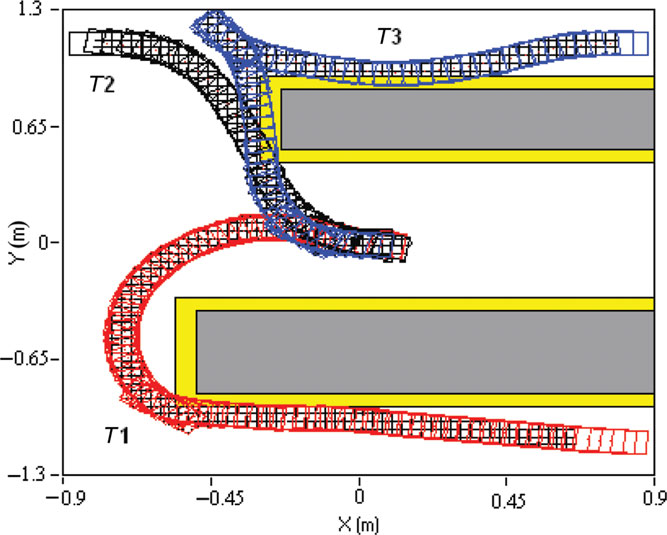}
      \caption{Path planning results from \cite{Gomez2012OptimalVehicles}}
\label{Fig:gomez}
\end{figure}

An other interesting example of classic Q-learning is described in \cite{Gomez2012OptimalVehicles} where the authors designed an agent for the path planning problem of a ground vehicle considering obstacles with Ackermann steering by using $(v,x,y,\theta)$ (speed, positions and heading) as state representation, and used reinforcement learning as an optimizer (See Fig. \ref{Fig:gomez}).

Though one would expect that machine learning could give an overall end-to-end solution to automated driving, the study of the recent literature shows that Reinforcement Learning research can give answers to certain sub-tasks of this problem. The papers in recent years can be organized around these problems, where a well-dedicated situation or scenario is chosen and examined whether a self-learning agent can solve it. These problem statements vary in complexity. As mentioned earlier, the complexity of reinforcement learning, and thus training time, is greatly influenced by the complexity of the problem chosen, the nature of the action space, and the timeliness and proper formulation of rewards. The simplest problems, such as lane-keeping or vehicle following, can generally be traced back to simple convex optimization or control problems. However, in these cases, the formulation of secondary control goals, such as passenger comfort, is more comfortable to articulate. At the other end of the imagined complexity scale, there are problems, like in the case of maneuvering in dense traffic, the efficient fulfillment of the task is hard to formulate, and the agent needs predictive "thinking" to achieve its goals. In the following, these approaches are presented.

\subsection{Car following}
Car following is the simplest task in this survey, where the problem is formulated as follows: There are two participants of the simulation, a leading and the following vehicle, both keeping their lateral positions in a lane, and the following vehicle adjusts its longitudinal speed to keep a safe following distance. The observation space consists of the $(v, dv, ds)$ tuple, representing agent speed, speed difference to the lead, and headway distance. The action is the acceleration command. Reward systems use the collision of the two vehicles as a failure naturally, while the performance of the agent is based on the jerk, TTC (time to collision)  \cite{Zhu2019SafeDriving}, or passenger comfort \cite{YeAutomatedEnvironment}. Another approach is shown in \cite{Zhu2018Human-likeLearning}, where the performance of the car following agent is evaluated against real-world measurement to achieve human-like behavior.
\subsection{Lane keeping}
Lane-keeping or trajectory following is still a simple control task, but contrary to car following, this problem focuses on lateral control. The observation space in these studies us two different approaches: One is the "ground truth" lateral position and angle of the vehicle in lane \cite{Sallab2016End-to-EndAssist, Lee2017AutonomousQ-learning, Ma2018ImprovedLearning}, while the second is the image of a front-facing camera \cite{Wolf2017LearningQ-Networks, Xu2018AutonomousTranslation, Li2019ReinforcementNotes}. Naturally, for image-based control, the agents use external simulators, TORCS, and GAZEBO/ROS in these cases. Reward systems almost always consider the distance from the center-line of the lane as an immediate reward. It is important to mention that these agents hardly consider vehicle dynamics, and surprisingly does not focus on joined longitudinal control.    

 \subsection{Merging}
\begin{figure}[]
      \centering
      \includegraphics[width=\linewidth]{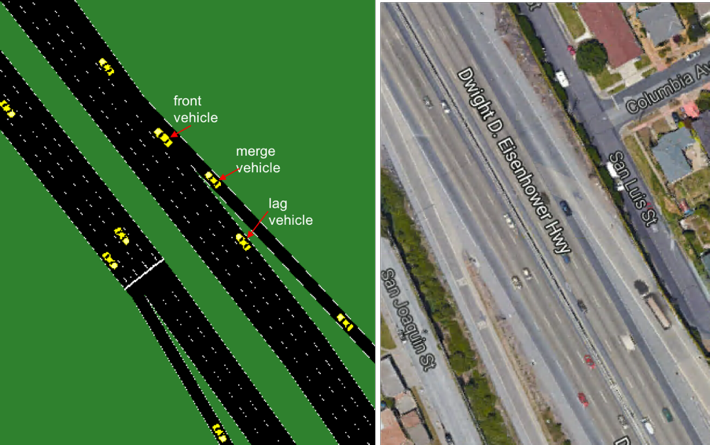}
      \caption{Ramp merge: (a) simulated scenario and (b) real-world location (source: \cite{Wang2017FormulationMerge})}
\label{Fig:merge}
\end{figure}

The ramp merge problem deals with the on-ramp highway scenario (see Fig. \ref{Fig:merge}), where the ego vehicle needs to find the acceptable gap between two vehicles to get on the highway. In the simplest approach, it is eligible to learn the longitudinal control, where the agent reaches this position, as can be seen in \cite{Wang2018AutonomousSpace, Wolf2018AdaptiveStates, Bouton2019Cooperation-AwareTraffic}. Other papers, like \cite{Wang2017FormulationMerge} use full steering and acceleration control. 
 In \cite{Wolf2018AdaptiveStates}, the actions control the longitudinal movement of the vehicle accelerate and decelerate, and while executing these actions, the ego vehicle keeps its lane. Actions "lane change left" as well as "lane change right" imply lateral movement. Only a single action is executed at a time, and actions are executed in their entirety, the vehicle is not able to prematurely abort an action. 
 
An exciting addition can be examined in \cite{Bouton2019Cooperation-AwareTraffic}, where the surrounding vehicles act differently, as there are cooperative and non-cooperative drivers among them. They trained their agents with the knowledge about cooperative behavior, and also compared the results with three differently built MTCS planners. Full information MCTS naturally outperforms RL, though they are computationally expensive. The authors used a curriculum learning approach to train the agent by gradually increasing traffic density. As they stated: "When training an RL agent in dense traffic directly, the policy converged to a suboptimal solution which is to stay still in the merge lane and does not leverage the cooperativeness of other drivers. Such a policy avoids collisions but fails at achieving the maneuver."

The most detailed description for this problem is given by \cite{Wang2017FormulationMerge}, where "the
driving environment is trained as an LSTM architecture to
incorporate the influence of historical and interactive driving behaviors on the action selection. The Deep Q-learning process takes the internal state from LSTM as the input to the Q-function approximator, using it for the action selection based on more past information. The Q-network parameters are updated with an experience replay, and a second target Q-network is used to relieve the problems of local optima and instability." With this approach, the researchers try to mix the possibilities from behavior prediction and learning, simultaneously achieving better performance. 

\subsection{Driving in traffic}

The most complicated scenario examined in the recent papers are those where the autonomous agent drives in traffic. Naturally, this task is also scalable by the topology of the network, the amount and behavior of the surrounding vehicles, the application of traffic rules, and many other properties. Therefore almost all of the current solutions deal with highway driving, where the scenario lacks intersections, pedestrians, and the traffic flow in one direction in all lanes. Sub-tasks of this scenario were examined in the previous sections, such as lane-keeping, or car following. In the following, two types of highway driving will be presented. First, the hierarchical approaches are outlined, where the agents act on the behavioral layer, making decisions about lane changing or overtaking and performs these actions with an underlying controller using classic control approaches. Secondly, end-to-end solutions are presented, where the agents directly control the vehicle by steering and acceleration.
As the problem gets more complicated, it is important to mention that the agents trained this would only be able to solve the type of situations that it is exposed to in the simulations. It is, therefore, crucial that the design of the simulated traffic environment covers the intended case \cite{Hoel2018AutomatedLearning}. 
 
Making decisions on the behavioral layer consists of at least three discrete actions: Keeping current lane, Change to the left, and Change to the right, as can be seen in \cite{Alizadeh2019AutomatedEnvironment}. In this paper, the authors used the ground truth information about the ego vehicle's speed and lane position, and the relative position and speed of the eight surrounding vehicles as the observation space. They trained and tested the agents in three categories of observation noises: noise-free, mid-level noise (\%5), and high-level noise (\%15), and showed that the training environments with higher noises resulted in more robust and reliable performance, also outperforming the rule-based MOBIL model, by using DQN with a DNN of ${64, 128, 128, 64}$ hidden layers with $tanh$ activation. 
In a quite similar environment and observation space, \cite{Hoel2018AutomatedLearning} used a widened set of actions to perform the lane changing with previous accelerations or target gap approaching, resulting in six different actions as can be seen in table \ref{tab:hoel}. They also achieved the result that the DQN agent - using two convolutional and one dense layer -    performed on par with, or better than, a reference model based on the IDM \cite{Treiber2000CongestedSimulations}. and MOBIL \cite{Kesting2007GeneralModels} model. In the other publication from the same author \cite{Hoel2019CombiningDriving}, the action space is changed slightly by changing the acceleration commands to increasing and decreasing the ACC set-points and let the underlying controller perform these actions.

\begin{table}[htb]
\centering
\caption{Action space in \cite{Hoel2018AutomatedLearning}}
\label{tab:hoel}
\begin{tabular}{l l}
\hline
$a_1$ & Stay in current lane, keep current speed \\
$a_2$ & Stay in current lane, accelerate with $-2 m/s^2$ \\
$a_3$ & Stay in current lane, accelerate with $-9 m/s^2$ \\
$a_4$ & Stay in current lane, accelerate with $2 m/s^2$ \\
$a_5$ & Change lanes to the left, keep current speed \\
$a_6$ & Change lanes to the right, keep current speed \\ \hline
\end{tabular}
\end{table}

In \cite{Shi2019DrivingLearning}, a two-lane scenario is considered to distribute the hierarchical decisions further. First, a DQN makes a binary decision about "to or not to change lane", and afterward, the other Q network is responsible for the longitudinal acceleration, based on the previous decision. Hence the second layer, integrated with classic control modules (e.g., Pure Pursuit Control), outputs appropriate control actions for adjusting its position.
In \cite{Xu2018AHighways}, the above mentioned two-lane scenario is considered, though the authors used an actor-critic like learning agent.

\begin{figure}[]
      \centering
      \includegraphics[width=\linewidth]{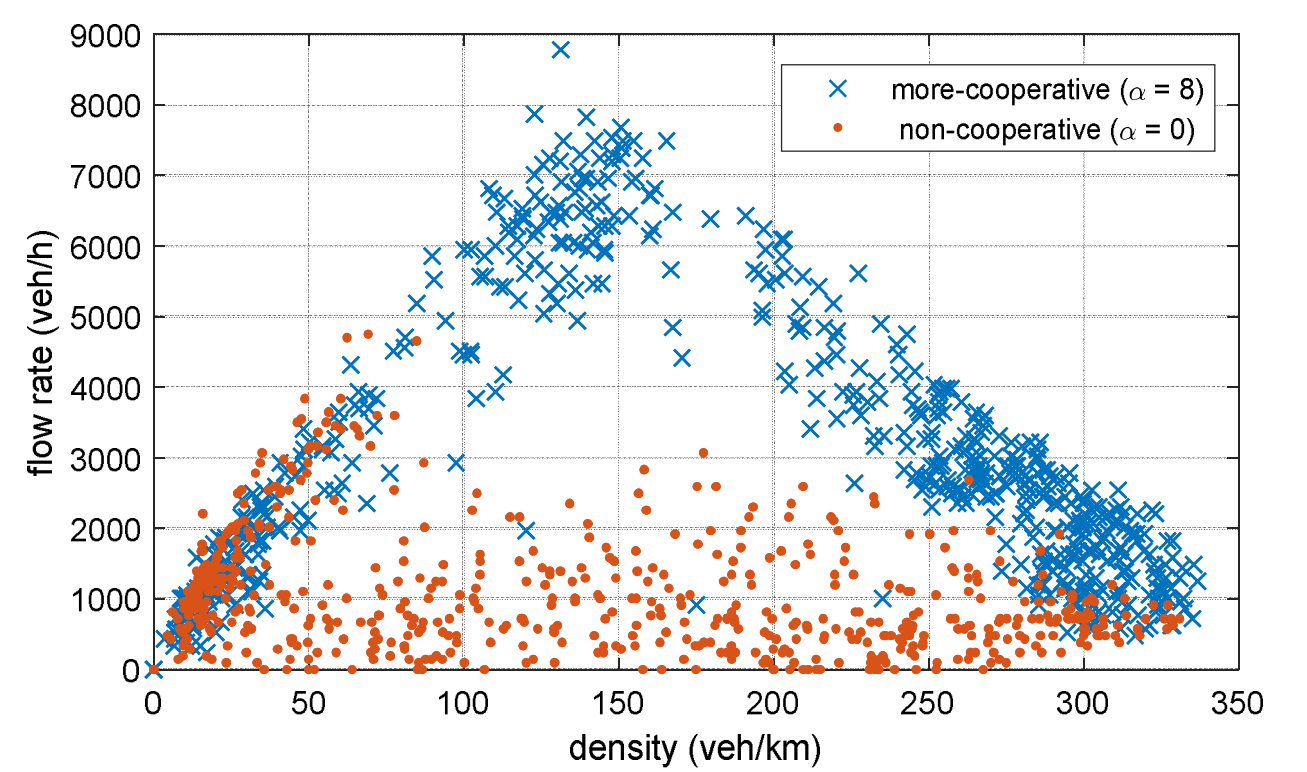}
      \caption{Flow-density relations detected by the virtual loops under different strategies (source:\cite{Wang2019CooperativeLearning})}
\label{Fig:Wang2019CooperativeLearning}
\end{figure}

An interesting question in automated driving is the cooperative behavior of the trained agent. In \cite {Wang2019CooperativeLearning} the authors considered a three-lane highway with a lane-based grid representation as observation space and a simple tuple of four for action space {left, right, speedup, none}, and used the reward function to achieve cooperative and non-cooperative behaviors. Not only the classic performance indicators of the ego vehicle is considered in the reward function, but also the speed of the surrounding traffic, which is naturally affected by the behavior of the agent. The underlying network uses two convolutional layers with 16 filters of patch size (2,2) and RELU activation, and two dense layers with 500 neurons each.  
To evaluate the effects of the cooperative behavior, the authors collected traffic data by virtual loops in the simulation and visualized the performance of the resulting traffic in the classic flow-density diagram (see Fig. \ref{Fig:Wang2019CooperativeLearning}.)
It is shown that the cooperative behavior results in higher traffic flow, hence better highway capacity and lower overall travel time.

The realism of the models could still differentiate end-to-end solutions. For example, in \cite{Bai2019DeepTraffic}, instead of using the nonholonomic Ackermann steering geometry, the authors use a holonomic robot model for the action space, which highly reduces the complexity of the control problem. Their actions are Acceleration, Deceleration, Change lane to the left, Change lane to the right, and Take no action, where the first two apply maximal acceleration and deceleration, while the two lane-changing actions simply use constant speed lateral movements. They use Dueling DQN and prioritized experience replay with a grid-based observation model.
A similar control method and nonholonomic kinematics is used in \cite{Nageshrao2019AutonomousLearning}. The importance of this research is that it considers safety aspects during the learning process. By using an MPC like safety check, the agent avoids actions that lead to a collision, which makes the training faster and more robust.      

Using nonholonomic kinematics needs acceleration and steering commands. In \cite{Becsi2018HighwayLearning, Aradi2018PolicyDriving}, the authors used a continuous observation space of the structured information of the surrounding vehicles and Policy-gradient RL structure to achieve end-to-end driving. Since the utilized method has discrete action-space, the steering and acceleration command needed to be quantized. The complexity of driving in traffic with an end-to-end solution can be well examined by the number of training episodes needed by the agent. While in simple lane-keeping scenarios, the agents finished the task in few hundreds of episodes, the agent used for these problems needed 300'000.         

\section{Future Challenges}

The recent achievements on the field showed that different deep reinforcement learning techniques could be effectively used for different levels of autonomous vehicles' motion planning problems, though many questions remain unanswered. The main advantage of these methods is that they can handle unstructured data such as raw or slightly pre-processed radar or camera-based image information. 

Though using neural networks and deep learning techniques as universal function-approximators in automotive systems poses several questions. As stated in \cite{Falcini2017DeepSoftware}, function development for automotive applications realized in electronic control units (ECUs) is subject to proprietary OEM norms and several international standards, such as Automotive SPICE (Software Process Improvement and Capability Determination) \cite{2015AutomotiveModel} and ISO 26262 \cite{2011ISOVocabulary}. However, these standards are still far from addressing deep learning with dedicated statements, since verification and validation is not a solved issue in this domain. Some papers deal with these issues by using an underlying safety layer, which verifies the safety of a planned trajectory before the vehicle control system executes it. However, full functional safety coverage can not be guaranteed in complex scenarios this way. 

One of the main benefits of using deep neural networks trained by a reinforcement learning agent in motion planning is the relatively low computational requirements of the trained network. Though this property needs a vast amount of trials in the learning phase to gain enough experience, as mentioned before, for simple convex optimization problems, the convergence of the process is fast. However, for complex scenarios, the training can quickly reach millions of steps, meaning that one setup of hyper-parameters or reward hypothesis can last hours or even days. Since complicated reinforcement learning tasks need continuous iteration on the environment design, network structure, reward scheme, or even the used algorithm itself, designing such a system is a time-consuming project. Besides the appropriate result analysis and inference, the evaluation time highly depends on the computational capacities assigned. On this basis, it is not a surprise that most papers nowadays deal with minor subtasks of the motion planning problem, and the most complex scenarios, such as navigating in urban traffic, can not be found in the literature.

By examining the observation element of the recent articles, it can be stated that most researches ignore complex sensor models. Some papers use "ground truth" environment representations or "ideal" sensor models, and only a few articles utilize sensor noise. On the one hand, transferring the knowledge acquired from ideal observations to real-world application poses several feasibility questions \cite{Szalay2018DevelopmentConsiderations}, on the other hand, using noisy or erroneous models could lead to actually more robust agents, as stated in \cite{Alizadeh2019AutomatedEnvironment}. 

The same applies to the environment, which can be examined best amongst the group of highway learners, where the road topology is almost always fixed, and the surrounding vehicles' behavior is limited. Validation of these agents is usually made in the same environment setup, which contradicts the basic techniques of machine learning, where the training and validation scenarios should differ in some aspects. As a reinforcement learning agent can generally act well in the situations that are close to those it has experience with, it is crucial to focus on developing more realistic and diverse environments, including the modeling level of any interacting traffic participant to achieve such agents that are easily transferable to real-world applications. This applies to vehicle dynamics, where more diverse and more realistic modeling would be needed. Naturally, these improvements increase the numerical complexity of the environment model, which is one of the main issues in these applications. 

Tending towards mixed or hierarchical system design would be a solution to this problem in the future, by mixing classic control approaches and deep RL. Also, the use of extended learning techniques, such as curriculum learning, transfer learning, or Alpha-Go like planning agents, would profoundly affect the efficiency of these projects. 

Overall it can be said that many problems need to be solved in this field, such as the detail of the environment and sensor modeling, the computational requirements, the transferability to real applications, robustness, and validation of the agents. Because of these issues, it is hard to predict whether reinforcement learning is an appropriate tool for automotive applications.

\section*{Acknowledgment}
The research reported in this paper was supported by the Higher Education Excellence Program of the Ministry of Human Capacities in the frame of Artificial Intelligence research area of Budapest University of Technology and Economics (BME FIKPMI/FM).

\bibliographystyle{IEEEtran}
\bibliography{MendeleyRefs}

% \begin{IEEEbiography}[{\includegraphics[width=1in,height=1.25in,clip,keepaspectratio]{Figures/Aradi_Szilard_small.jpg}}]{Szilárd Aradi}
%  received the M.Sc. degree in 2005 and Ph.D. in 2015 from the Budapest University of Technology and Economics, Budapest, Hungary, where he is currently working with the Department of Control for Transportation and Vehicle Systems. Since 2016, he has been an Senior Lecturer at the Department of Control for Transportation and Vehicle Systems, Budapest University of Technology and Economics. His research interests include embedded systems, communication networks, vehicle mechatronics, and reinforcement learning. His research and industrial works have involved railway information systems, vehicle on-board networks, and vehicle control.
% \end{IEEEbiography}

\end{document}